\documentclass[sigconf]{aamas} 

\usepackage{balance} 
\usepackage{tabularx}
\usepackage{multirow}

\setcopyright{ifaamas}
\acmConference[AAMAS '24]{Proc.\@ of the 23nd International Conference
on Autonomous Agents and Multiagent Systems (AAMAS 2024)}{May 29 -- June 2, 2024}
{London, United Kingdom}{A.~Ricci, W.~Yeoh, N.~Agmon, B.~An (eds.)}
\copyrightyear{2024}
\acmYear{2024}
\acmDOI{}
\acmPrice{}
\acmISBN{}

\title{PPS-QMIX:Periodically Parameter Sharing for Accelerating Convergence of Multi-Agent Reinforcement Learning}

\author{Ke Zhang}
\affiliation{
  \institution{China University of Petroleum}
  \city{Beijing}
  \country{China}}
\email{2021211702@student.cup.edu.cn}

\author{DanDan Zhu}
\affiliation{
  \institution{China University of Petroleum}
  \city{Beijing}
  \country{China}}
\email{zhu.dd@cup.edu.cn}

\author{Qiuhan Xu}
\affiliation{
  \institution{China University of Petroleum}
  \city{Beijing}
  \country{China}}
\email{xu_qiuhan@qq.com}

\author{Hao Zhou}
\affiliation{
  \institution{China University of Petroleum}
  \city{Beijing}
  \country{China}}
\email{zhouh68@foxmail.com}

\author{Ce Zheng}
\affiliation{
  \institution{Télécom Paris}
  \city{Paris}
  \country{France}}
\email{ce.zheng@telecom-paris.fr}
\begin{abstract}
Training for multi-agent reinforcement learning(MARL) is a time-consuming process caused by distribution shift of each agent. One drawback is that strategy of each agent in MARL is independent but actually in cooperation. Thus, a vertical issue in multi-agent reinforcement learning is how to efficiently accelerate training process. To address this problem, current research has leveraged a centralized function(CF) across multiple agents to learn contribution of the team reward for each agent. However, CF based methods introduce joint error from other agents in estimation of value network. In so doing, inspired by federated learning, we propose three simple novel approaches called {\romannumeral1}:Average Periodically Parameter Sharing(A-PPS), {\romannumeral2}:Reward-Scalability Periodically Parameter Sharing(RS-PPS) and {\romannumeral3}:Partial Personalized Periodically Parameter Sharing(PP-PPS) mechanism to accelerate training of MARL. Agents share Q-value network periodically during the training process. Agents which has same identity adapt collected reward as scalability and update partial neural network during period to share different parameters. We apply our approaches in classical MARL method QMIX and evaluate our approaches on various tasks in StarCraft Multi-Agent Challenge(SMAC) environment. Performance of numerical experiments yield enormous enhancement, with an average improvement of 10\%-30\%, and enable to win tasks that QMIX cannot. Our code can be downloaded from \href{https://github.com/ColaZhang22/PPS-QMIX}{https://github.com/ColaZhang22/PPS-QMIX}. 
\end{abstract}


\newcommand{\BibTeX}{\rm B\kern-.05em{\sc i\kern-.025em b}\kern-.08em\TeX}

\begin{document}


\pagestyle{fancy}
\fancyhead{}


\maketitle 


\section{Introduction}
In recent years, Multi-agent reinforcement learning(MARL) has recently obtained enormous success for handling various optimization tasks, ranging from electric games\cite{DBLP:journals/nature/VinyalsBCMDCCPE19,DBLP:books/ox/22/WangWEPTK22,DBLP:conf/nips/TerryBGJHSSDHPW21}, autonomous driving\cite{DBLP:conf/aaai/ZhaoWXCL0L22,DBLP:journals/eswa/SilvaSSB19}. Compared with reinforcement learning which just has one agent, MARL has to tackle several challenges. Due to decentralized value function or policy and limitation of communication, one issue is that algorithm also has high instability and deviation caused by insufficient estimation and sampling. Another issue is curse of dimensionality caused by large dimensional observation and action spaces. A common way\cite{DBLP:conf/nips/LoweWTHAM17,DBLP:conf/nips/YuVVGWBW22} to address these problems is referred as centralized training and decentralized execution(CTDE)\cite{DBLP:conf/aaai/FoersterFANW18}, has been adapted in most MARL algorithms, which all agents share only one parameterized function for all policy network or Q-value function to estimate contribution of each agent. Recent approaches 
VDN\cite{DBLP:conf/atal/SunehagLGCZJLSL18}, QTran\cite{DBLP:conf/icml/SonKKHY19}and QMIX~\cite{DBLP:conf/icml/RashidSWFFW18} focus on building an heuristic appropriate function to measure the proportion of each agent's contribution to the global reward. 

However, in real world circumstance, like autonomous vehicles\cite{DBLP:journals/tii/CaoY0C13} and internet of things devices\cite{DBLP:journals/iotj/MohammadiAGO18}, privacy of sensitive data is a crucial issue that consists of surrounding sensitive state information which cannot share with other vehicles or devices. Thus, we consider a decentralized training decentralized execution(DTDE) circumstance that each agent can just visit it own experience buffer but cannot access to other experience trajectory. The other drawback is this situation comes with a vital challenge limited access of local information efficiently increase fluctuations of exploration for each agent. As shown in fig, due to the difference of policy, exploration of each agent enables to be regarded as a non-independent and identically distribution(Non-IID)\cite{DBLP:conf/icde/LiDCH22}. Therefore, for each agent, distribution drift in exploration leads policy falls into sub-optimal point and ignores cooperative relationship with other agent.
\begin{figure}[htbp]
    \centering
    \includegraphics[width=0.45\textwidth]{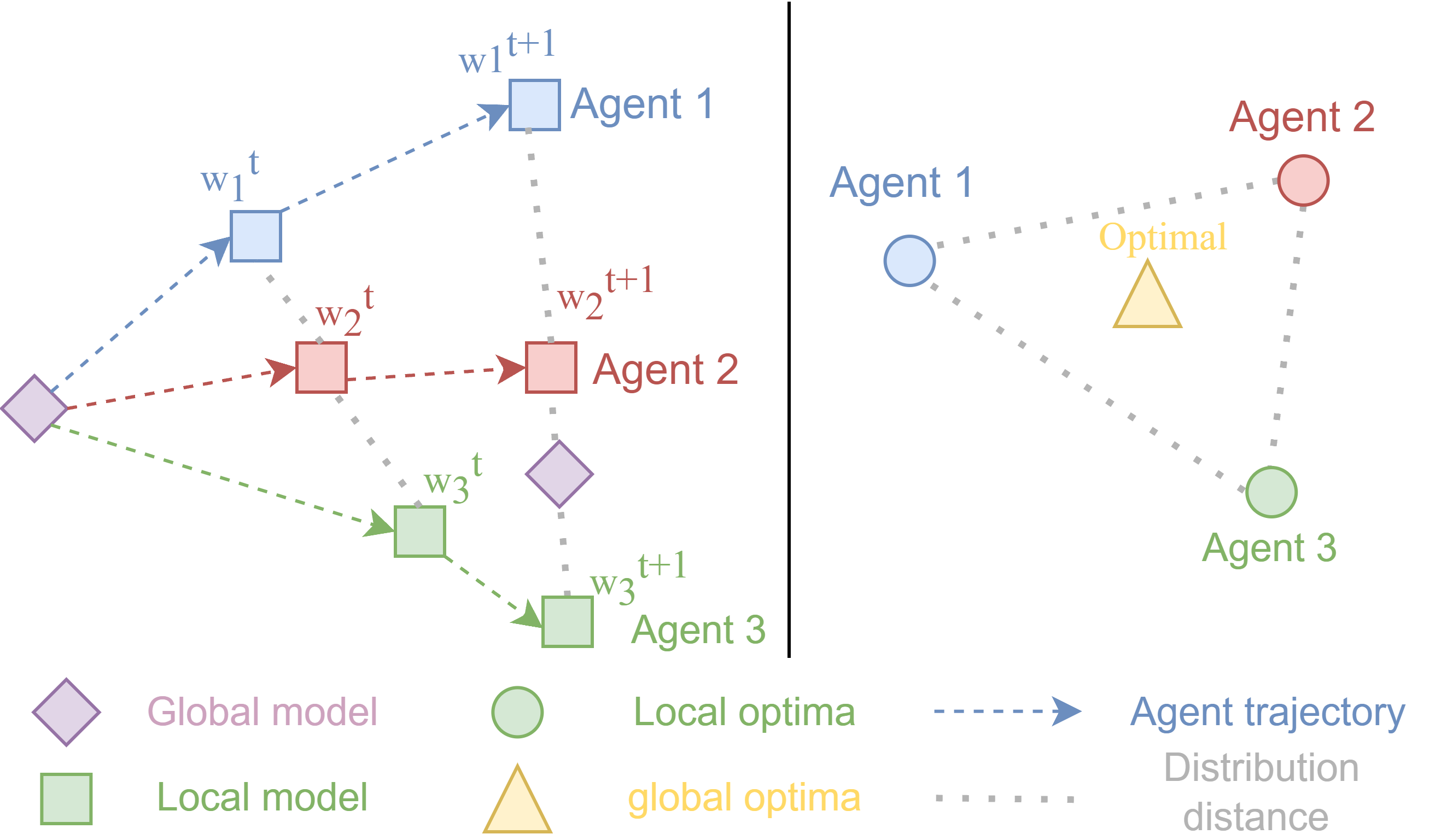}
        \caption{Distribution drift in decentralized training for agents. Due to distribution drift in exploration process, each agent enable to acquire local optima but cannot get to global optima. To solve Non-IID in agent experience trajectory, each agent explore their own environment and transmitted local model to aggregate into a generalized model.}
    \label{fig:issue}
\end{figure}
\par
Motivated by these challenges, inspired by concept of federated learning\cite{DBLP:journals/iotj/WuUHKSV22,DBLP:conf/iotdi/YuCVZEZMR23}. We first propose average periodically parameter sharing(A-PPS), which means each agent shares an average weight in aggregation phase. 
However,in some tasks, due to difference of exploration, A-PPS cannot appropriately reflect effectiveness in their process of exploration cause some agents do not acquire useful experience. Instead of taking the average of model weights, a reward buffer is used as model weights to measure the quality of agents exploration process. Besides, proved by personalized federated learning, for each agent, part of their own model can be seen as combination between personalized feature representation and value representation. We just aggregate value representation part of module and allow agent to have personalized feature representation part. Thus, based on naive FedAvg methods, we propose three simple novel approaches called average periodically parameter sharing(A-PPS), reward-scalability periodically parameter sharing(RS-PPS) and Partial Personalized Periodically Parameter Sharing(PP-PPS). 
\par
Conclusively, we evaluate A-PPS, RS-PPS and PP-PPS in multiple scenarios from SMAC simulated environment, a benchmark of multi-agent cooperation environment.  A-PPS, RS-PPS and PP-PPS bring obvious enhancement compared with QMIX and VDN in different tasks of SMAC environment\cite{DBLP:conf/atal/SamvelyanRWFNRH19}. In some environments, in comparison with classical A-PPS, our approaches possess faster convergence rate and has a increasing efficiency. Notably, in some asymmetric competition task like corridor, our RS-PPS and PP-PPS has a increasing privacy and effectiveness. Our source code is available from URL: \url{https://www.acm.org/publications/proceedings-template}. The following is a list of the contributions of this paper: 
\begin{enumerate}
    \item Inspire by FedAvg, we introduce federated learning in classical mutli-agent reinforcement learning approach QMIX and perform a average periodically parameter sharing(A-PPS) and evaluate A-PPS into SMAC to prove its validation and efficiency.
    \item  Due to distribution drift in multi-agent exploration trajectory, we propose a novel approach called reward-scalability periodically parameter sharing(RS-PPS) which use a accumulated reward function to measure weight of aggregation for each agent. 
    \item Inspired by modification about personalized federated learning\cite{DBLP:conf/iclr/OzkaraGDD23,DBLP:conf/icml/Dai0H0T22}, we divided value network into two parts, which one is feature representation and the other is value network, to increase learning adaptation and keep individual personality for multi-agent.
\end{enumerate}
\begin{figure*}[t]
    \centering
    \includegraphics[width=0.8\textwidth]{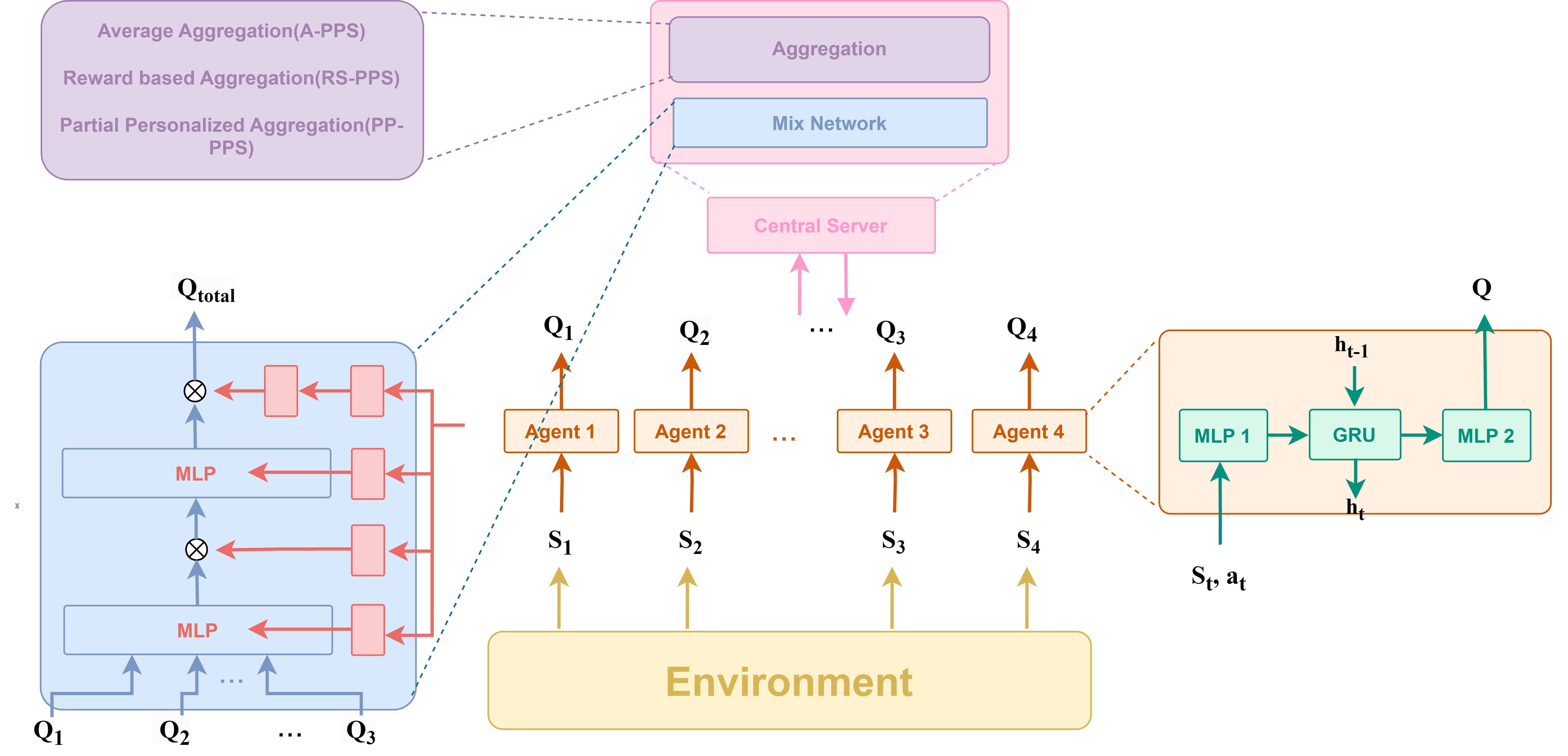}
        \caption{Architecture of parameter sharing QMIX. Middle of figure represents overall architecture. Each agent in architecture has its own value network like brown component in right part. The purple components are aggregation block. Value network parameters of each agent are shared by three approaches periodically to enhance experience of each agent.}
    \label{fig:architecture}
\end{figure*}
\section{Background}
Multi-agent reinforcement learning(MARL)\cite{DBLP:conf/nips/FoersterAFW16} consist of multiple agents can make decisions cooperatively in a shared environment. Unlike traditional distributed single-agent reinforcement learning, agents in MARL do not exist independently but in cooperation or competition relationships. Multi agent reinforcement learning can be formalized as a decentralized partially observable Markov decision process (Dec POMDP) problem\cite{DBLP:series/sbis/OliehoekA16}. Dec-POMDP generally consists of multiple tuple (\textit{$S_t$,$A_t$,$R_t$,$\gamma_t$}). $S_t=(s^1_t,s^2_t,...,s^n_t)$ is joint state and $A_t=(a^1_t,a^2_t,...,a^n_t)$ represents joint actions adapted by agent at time t and n represent numbers of agent in common environment, the whole process of MARL can be described in Fig.\ref{Framework of Multi-Agent Reinforcement Learning}. Compare with RL, in multi-agent reinforcement learning, the state transition functions of agents in the same state will still change due to the effect of the actions from other agents and thus do not satisfy the time invariance of the Markov process, which means for each agent in environment:
$$ P(s_{t^{'}}|s_t,a_t) \ne P(s_{t^{1'}}|s_{t^1},a_{t^1})  \ \ \  with \ \  s_t=s_{t^1}$$
To address this problem, recent studies can be developed into various categories including value-based and policy-based approaches. The first employs one or more centralized functions (CFs) to comprehend the impact of the agents’ actions to the team goal, such as VDN, QMIX. The CFs allow to optimize the agents’ parameters with respect to a global team reward. 
\par
At the same time, in the process of exploring and gaining experience, due to the initial random strategy, some agents take the correct action when facing different scenarios of cooperative tasks, while others take the wrong action. In the CF training paradigm, these correct and incorrect behaviors are treated as a whole and trained as a sample. Therefore, it will increase the exploration space and lead to inconsistent iteration directions of the intelligent agent value network during the training process, resulting in convergence difficulties and falling into local optima. In this situation, intelligent agents are unable to maximize the global value function and complete collaborative tasks. Therefore, another major challenge in solving multi-agent problems is how to ensure correct information and share it with other agents during differentiated exploration processes.
\section{Related Work}
\subsection{Federated Learning}
Distribution of data is a key component in federated learning\cite{DBLP:conf/atal/HebertGPC23}. The pervasiveness of multiple agents in real world like vehicles, has led to the rapid growth of private data originating from distributed sources. Meanwhile, distribution of trajectory in multi-agent reinforcement learning plays a significant role and influences efficiency of training. Both distribution of data and trajectory in multi-agent environment is a non-independent and identically distribution(Non-IID)\cite{DBLP:conf/nips/WangLLJP20}, which each agent learns deviant knowledge from different experience and generates a preferred strategy so that falls into local optima like in \ref{fig:issue}.
\par
To overcome Non-IID constraints and protect privacy, FedAvg has been proposed by \cite{DBLP:conf/aistats/McMahanMRHA17} to solve unbalanced data distribution and protect privacy. FedAvg is a promising method to train the neural network parameters while keeping the training data in the local devices so that can protect sensitive data. For each agent in MARL, given a learning 

A typical process of FedAvg with k agents can formulize as follow and average gradient on its local data at the current model $w_t$, For federated multi-agent reinforcement learning, given a set of $N$ agents with loss functions ${F_i}_{i=1}^N$ and datasets distribution ${D_i}_{i=1}^N$ interested in formulating a joint cooperative model, FedAvg updates a centralized model as:
$$For \ \ each \ \ agent \ \ i:w_i(t+1)=w_i(t)-\gamma \nabla F_i(w_i(t))$$
$$If \ \ t \ \ mod \ \ interval=0: \ \ w_g(t+1)=\sum_{i=1}^{N}\frac{|D_i|}{|D|}w_i(t+1)$$
where $w_i(t)$ is the model update computed by agent $i$ at time step $t$, $\gamma$ is a fixed learning rate and $w_g$ are the 3parameters of the centralized model and will be synchronized periodically into each individual agent. $D_i$ and $D$ denote datasets in each agent. $interval$ denotes the span between aggregations, after a span, agent learning novel knowledge and experience from trajectory. Meanwhile, how to determine weight for each agent in the environment is a critical issue. In traditional federated learning, due to the uneven distribution of the dataset, aggregation models can introduce new knowledge that is unknown to the current model. However, distribution of agent experience is Non-IID cause each agent has a different trajectory.  To dissipate negative impact of Non-IID datasets, some modified Federated learning approaches proposed that adjust weight of each client by distance and dataset amount. Compared with deep learning, reinforcement learning has a exclusive criteria to estimate efficiency of each agent module for MARL. Inspired by these modified approaches, reward is utilized as a criteria to aggregate the module. 
\par
Therefore, in this work, we introduce reward-based weight to modified and utilize FL into multi-agent reinforcement learning to accelerate training process. Instead of AvgFed, our approach measure and determine the weight based on accumulated reward for each agent in trajectory. We merge our approaches into classical value factorization methods QMIX for MARL and replace QMIX with a application of A-PPS, RS-PPS, PP-PPS. These approaches can be merged into various value factorization methods. Modified QMIX(A-PPS,RP-PPS,PP-PPS) has a significant improvement and benefits from other methods in different tasks in SMAC.

\subsection{Value Function Factorization}
A naive approach proposed\cite{DBLP:conf/iclr/0001WZZ20} by method to solve multi-agent reinforcement learning which concatenate and merge states and actions of all agents into an huge action and state space has been proved that fail to find a global optimum due to some lazy agents. This approach ignore individual max reward so that in some scenarios agents become lazy and cannot learn to cooperate with other agents to get a global optimum. Therefore, the key in MARL is to balance individual global-max(IGM). One method to address this issue by determining role and individual contribution for each agent is called Value Function Factorization. Current Value Function Factorization methods\cite{DBLP:conf/atal/HuangLSZLWMHWD22} focus on employing of a centralized function (CF) that learns each agent’s contribution to the team reward. 
\par
Classical Value function factorization method VDN consider a fully cooperative multi-agent reinforcement learning which each agent just enable to observe its own state and take action. In VDN, each agent possess a local $Q_i$ value network which is a part of global $Q_total$. VDN assume the global value function $Q_total$ can be decomposed into individual value function $Q_i$ for each agent $i$:
$$arg \ \max_{a}Q_{total}(\textbf{s},\textbf{a}) =f\begin{pmatrix}arg \ \max_{a_1} Q_1(s_1,a_1) \\arg \ \max_{a_2} Q_2(s_2,a_2)  \\ ...\\arg \ \max_{a_i} Q_i(s_i,a_i) \end{pmatrix}$$
in which \textbf{s} is joint state of observation and \textbf{s} is joint action taken for all agents. Therefore, each value function $Q_i$ based observation for one agent can be summed by a centralized function. For VDN, $Q_total$ is the sum of the $Q_i$ values of each agent:
$$Q_{total}(\textbf{s},\textbf{a})=\sum_{i=1}^{N} Q_i(s_i,a_i)$$
QMIX propose a mixed network to modifies VDN by replacing CF in VDN, simple sum of $Q_i$ value, by a neural network and add restriction to keep monotonic of CT:
$$ Q_{total}(\textbf{s},\textbf{a})=R(Q_1(s_1,a_1),Q_2(s_2,a_2),...,Q_i(s_i,a_i),\textbf{s})$$

$$w_{t+1}=w_{t}+\nabla_w(Q(s_t,a_t)-(\gamma*r_t + Q(s_{t+1},a_{t+1},w_t)))$$
$$ for \ \  each \ \ agent \ \ i: \frac{\partial Q_{total}}{\partial Q_i}\ge 0 \ \ i\in\left \{ 1,...,N \right \}$$
In which $R(*)$ is a RNN neural network to estimate the contribution for each agent. Q-mix demonstrated enforcing positive weights on the mixer network is sufficient to guarantee improvement compared with VDN. There exist several modified QMIX method, such as TransfQMIX\cite{DBLP:conf/atal/GalliciMM23,DBLP:conf/icml/ParisottoSRPGJJ20}, Qtran\cite{DBLP:conf/icml/SonKKHY19}, OWQMix and CWQMix\cite{DBLP:conf/nips/RashidFPW20} to enhance learning efficiency. TransfQMix introduces multi-head attention mechanism\cite{DBLP:conf/nips/VaswaniSPUJGKP17} to deal with high dimensional state space so that learn a useful representation from global state $\textbf{s}$. 
$$Q_{total}(\textbf{s},\textbf{a})\longrightarrow Q_{total}(Attention(\textbf{s}),\textbf{a})$$
OWQMix and CWQmix introduce a weight function $w$ to estimate importance of each joint action.
\par 
However, previous approaches ignores a crucial component, imbalance learning trajectory for each agent in the process of training. Due to exploration in beginning step, each agent acquire a imbalance trajectories and get various experience from these trajectories. So each agent cannot finish cooperative task due to difference of experience.  Therefore, federated learning is introduced into our method aims to lead agent to learn experience from other agents.
\begin{figure*}[ht]
    \centering
    \includegraphics[width=1\textwidth]{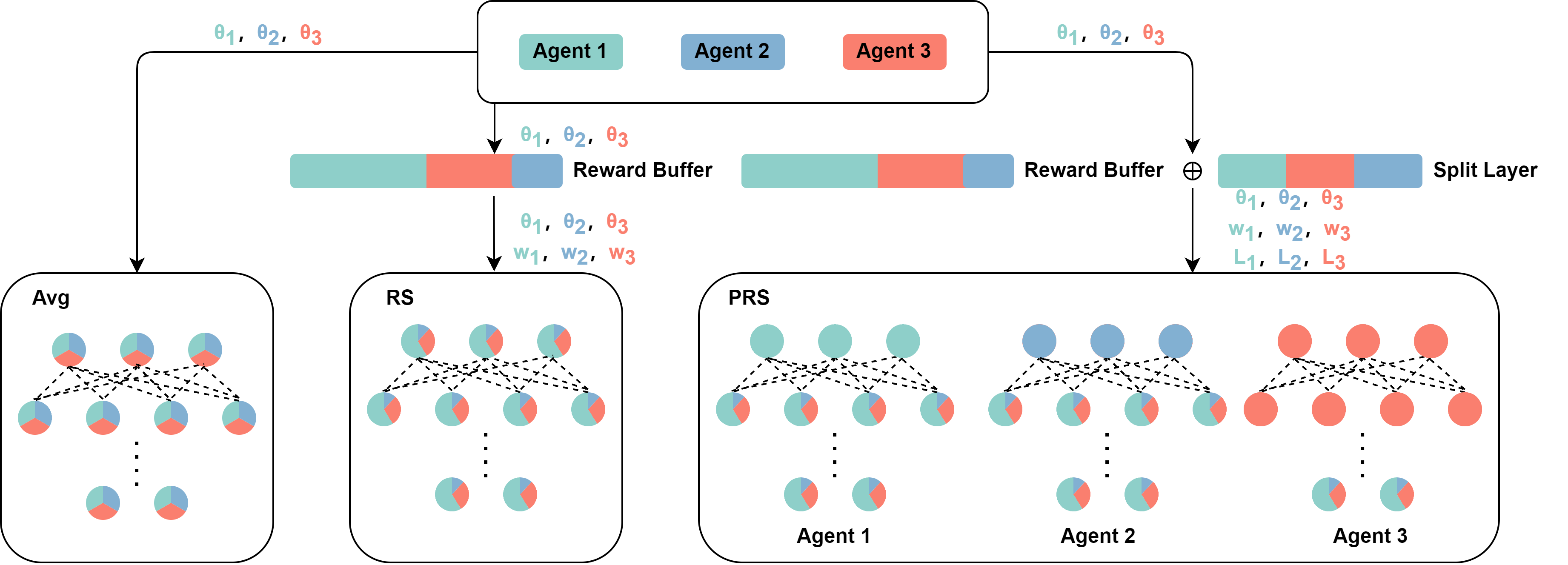}
        \caption{Three periodically parameters sharing approaches in QMIX. (a) Average periodically parameter sharing(A-PPS) adapts equal weight for each agent value network; (b) Reward-scalability periodically parameter sharing(RS-PPS) introduce a reward buffer to storage acquired reward in process of exploration as aggregate weight.
        (c) Partial Personalized Periodically Parameter Sharing(PP-PPS) divide agent value network into two parts, personalized representation and value function, PP-PPS keep personalized representation unchanged and just aggregates value function part.}
    \label{fig:methods}
\end{figure*}
\section{Method}
We consider a multi-agent cooperation problem where given $N$ agents, each agent can only access its own observations $s_i$ and generate action $a_i$ based on its own value function. In the process of multi-agent exploration of the environment, due to the large state space and random action strategies, the empirical trajectories obtained by each agent are biased and uneven. Sometimes, action taken by agent in observation is right but has no reward cause by other agent negative impact. Therefore, model training requires a lot of time to converge to the optimal value and achieve the optimal strategy. Meanwhile, due to limitations in access rights and privacy protection, intelligent agents are unable to learn to cooperate with other agents, resulting in local sub-optimal solutions. One way to solve this problem is to obtain parameter sharing through agents, enabling them to learn and collaborate with other agents to complete tasks. Inspired by federated learning, we propose the Average Parameter Sharing method based on the QMIX model to eliminate negative impact of imbalance experience between agents and accelerate training for convergence. The overall architecture for our method is shown as in figure \ref{fig:architecture}.  

\subsection{Average Periodically Parameter Sharing}
In order to overcome the imbalanced experience during the exploration process, inspired by FedAvg and for simplicity, our paper first proposes to average the weights of each agent into an entity and distribute them to each agent, as shown in fig \ref{fig:methods} (a). Because every intelligent agent learns different knowledge during the process of exploring the environment and gaining experience, simply iterating through their own experience and rewards can easily lead to incorrect iteration directions and local optima. The method of weight aggregation can transfer new knowledge learned by intelligent agents to other agents through weight sharing, avoiding falling into local optima and avoiding the leakage of private data. For the A-PPS method, the importance of each agent in MARL is equal.  A-PPS enable to be summarized as follow:
$$w_g(t+1)=\sum_{i=1}^{N}\frac{1}{N}w_i(t+1)$$
in which $w_g$ denotes global parameters of value network and $w_i$ denotes local parameters for each value networks. In the process of exploring the environment, as the number of agents in the task and environment increases, the observation state dimension of the agent explodes. Small parameter differences in the value function of intelligent agents can lead to unstable action selection, resulting in a huge deviation from the optimal distribution. Therefore, average parameter sharing is not suitable in this situation, and it cannot guarantee the acquisition of effective knowledge from other intelligent agents during the aggregation process. Therefore, it is necessary to propose a standard to measure the knowledge learned by intelligent agents during the training process.
\subsection{Reward-Scalability Periodically Parameter
Sharing}
In order to solve the deviation problem caused by the high-dimensional state space problem mentioned above, fed prox in Non IID datasets is used to reduce the convergence problem caused by the deviation of exploration and experience acquisition. FedProx\cite{DBLP:conf/mlsys/LiSZSTS20} is an improved federated learning algorithm used to address non independent identically distributed (non IID) data and device heterogeneity issues in federated learning. The distribution and amount of data in FedProx are used to determine the weights occupied by each model during the aggregation process. In multi-agent reinforcement learning, due to the cooperative nature of tasks, at time t, one agent takes the correct behavior, while other agents take the incorrect behavior. Therefore, there is also a problem of non independent and identically distributed multi-agent exploration. To address this issue, our method utilizes the rewards obtained by agents during the exploration process to measure the differences in distribution. Therefore, a novel method called  Reward-scalability periodically parameter sharing(RS-PPS) in fig \ref{fig:methods} (b) is proposed and a reward buffer is proposed to evaluate the differences in experience obtained by different agents during exploration:
$$reward \ \ buffer \ \ for \ \ agent \ \ i: \textbf{R}_i=\sum_{t=1}^{T}r_t  $$
$$w_g(t+1)=\sum_{i=1}^{N}\frac{R_i}{\sum_{i=1}^{N}R_i}w_i(t+1)$$
where $R_i$ denotes accumulate reward during time interval $T$ for agent $i$ and is used to estimate quality of exploration of agent. Compared to A-PPS, PS-PPS can adjust aggregation weights based on the rewards obtained by the agent during the exploration process. The more reward functions obtained during the exploration process, the closer the agent is considered to be to the global optimum, and therefore will be given higher weights. This process can be seen as imparting new experiential knowledge to other agents, while also avoiding the tendency towards the agent receiving the most rewards and falling into suboptimal solutions.
\subsection{Partial Personalized Periodically Parameter Sharing}
Characterized agents are a very effective way to solve collaborative tasks, where agents play different roles in the team to complete their own tasks and maximize the global reward function. In collaborative tasks, determining one's own role in the team is a very challenging issue, as correctly characterized agents can complete tasks within the team. Therefore, we propose a Partial Personalized Periodic Parameter Sharing (PP-PPS) in fig \ref{fig:methods} (c) method that divides the neural network of agent into two parts. The first part is the agent characterization network, where agents with different roles will focus on different parts of the environment. The second part is the intelligent agent value network, which evaluates the value of state action pairs in the environment. Therefore, during the aggregation process, only the second part of the network needs to be aggregated:
$$
\begin{cases}
w_{i,j}(t+1)= w_{i,j}(t+1)  &  \text{ if } j \ \ not \ \ in \ \ layer  \\
w_{i,j}(t+1)=\sum_{i=1}^{N}\frac{R_i}{\sum_{i=1}^{N}R_i}w_{i,j}(t+1)  & \text{ if } j \ \ in \ \ layer
\end{cases}
$$
in which $j$ donates layer in neural network of QMIX for each agent. Currently, a fixed point is used for the boundary point part of a two-part network. When parameters $w_{i,j}$ are located in the characterization network layer, they will not participate in parameter sharing to ensure the characterization of the intelligent agent. However, when the parameters $w_{i,j}$ are located in the value network part of the agent, the agent will join parameter sharing to accelerate the convergence of the network.The role of intelligent agents in collaborative tasks varies, so the required representation networks are also different. Due to the instability of the network model caused by fixed segmentation points, the convergence speed of the PP-PPS method is not as fast as the first two methods in most tasks.
\section{Experiment}
These three approaches, A-PPS, RS-PPS and PRS-PPS is implemented based on Python and Pytorch package. And the all experiments were run in parallel on a computing cluster consisting of two RTX 3080 GPU, 24 GB memory. In each task, same hyper-parameters in table 1 are adapted in our experiment, with exception that we train for 1e6 or 2e6 training iterations caused by computation restriction.
\par
To verify effectiveness of three proposed approaches, we selected QMIX as  basic model and evaluated proposed approaches in a SMAC environment. Then meanwhile we compared results of these methods with conventional value decomposition approaches QMIX and VDN in different SMAC tasks.
\par
Modified QMIX architecture described in section 3 is employed for A-PPS,RS-PPS, PRS-PPS. For each task, we fixed the random number seed, and evaluated 32 times every 5000 steps to obtain the winning rate. After training a certain number of times, we believe that the agent is believed to learned different tendency novel knowledge and therefore lead agent shared it with other agents through different methods to accelerate training and complete collaborative tasks. For A-PPS method, each agent shares knowledge equally with other agents. RS-PPS method uses a reward buffer of the same length as the batch size to evaluate the effective knowledge learned by agent during the training process, and aggregates it based on the weight of a reward buffer shown in section 4.3. In final PP-PPS method, a portion of the value network is selected as a personalized representation layer, while the other portion is regarded as value representation layer and aggregated through the RS-PPS method. The experimental results indicate that for conventional QMIX methods, our methods can accelerate network training and enable agents to complete tasks that cannot be completed by conventional QMIX methods.
\subsection{SMAC}
The SMAC environment, a micro unit control game environment based on electric game,StarCraft 2, was used to test the three methods we proposed. In SMAC environment, multiple agents need to collaborate against and beat enemies to win the game. Each agent has an limited observation range and can only observe the status of friendly forces within the range. The observed feature vectors consist of attribute information of friends and enemies: [distance, relative x, relative y, health, shield, unit type]. Each agent only enable to access their own observation state and rewards obtained during the exploration process. 
\par
In every exploration step, agent just receives local information from its perspective. There is no difference between friendly units in the field and dead friendly units from the perspective of the intelligent agent. According to the number of enemies and the number of our intelligent agents, tasks can be divided into asymmetric tasks, which the number of allied agents varies from enemies, and symmetric tasks ,which the number of allied agents is same as enemies. Finally, we tested our proposed algorithm in both circumstance: 3m, 2s3z and asymmetric adversarial tasks: 10m vs 11m, MMM2, Corridor and compared it with QMIX and VDN algorithms.
A-PPS, RS-PPS and PP-PPS can be directly applied in conventional MARL algorithm and do not need to add other architecture.
\begin{table}
  \caption{Hyper-parameters Used}
  \label{tab:locations}
 \begin{tabularx}{0.95\hsize}{X|X}
 \toprule
    \textbf{Parameter} & \textbf{Value} \\ 
    \midrule
    Max Train Steps & 1e6 \\
    Evaluate Freq & 5000 \\
    Target Update Freq & 200 \\
    Base Algorithm & VDN, QMIX \\
    Epsilon Decay Steps & 50000  \\
    Epsilon Max & 1  \\ 
    Epsilon Min & 0.05  \\ 
    Buffer Size & 5000 \\
    Batch Size & 96 \\
    Learning rate & 5e-4 \\
    Gamma & 0.99 \\
    Mixed Hidden Num & 1 \\
    Mixed Hidden Dim & (State Dim,64) \\ 
    Optimizer &   Adam   \\
    Grad Clip & True \\
    Activation Function & ReLu \\
    Orthogonal Initialization & True\\
    Lr Decay & False \\
    \midrule
    \textbf{Parameter sharing} & \\
    \midrule
    Soft Update & 0.05 \\
    Reward Buffer Size & 96 \\ 
    Aggregation Freq & 300 \\
    Personalized Layer & 4 \\
    \bottomrule
\end{tabularx}
\end{table}
\section{Result}
As shown in Figure \ref{fig:result}, our method performs only slightly faster than QMIX and VND in SMAC 3m and 2s3z tasks after 1 million training steps. We speculate that due to the low number of agents in the 3m (total of six agents) and 2s3z (total of ten agents) maps, the state space dimension is low and the distribution deviation caused by exploration are relatively small. Therefore, the effectiveness of our method is not significant and may even amplify the erroneous experience. At the same time, both 3m and 2s3z are symmetric adversarial tasks, with the same number and types of intelligent agents on both sides. Therefore, there is no significant difference in local observation of intelligent agents, resulting in relatively few erroneous experiences and a relatively average distribution. This leads to the best performance of A-PPS in our method. For instance, for a certain intelligent agent in 3m, only 12 situations need to be considered when there are 0, 1, 2 allies and 0, 1, 2, 3 enemies in the field of view. Therefore, the observation space of intelligent agents is relatively small.
\begin{table}
  \caption{Main Result}
  \label{tab:locations}
 \begin{tabularx}{1\hsize}{c|ccccc}
 \toprule
   \textbf{Scenarios}& \textbf{A-PPS} & \textbf{RS-PPS} & \textbf{PP-PPS} & \textbf{QMIX} & \textbf{VDN} \\ 
    \midrule
    3M & \textbf{0.96} & 0.91 & 0.83 & 0.87 & 0.95 \\
     2s3z & \textbf{0.85} & 0.82 & 0.71 & 0.73 & 0.77 \\
    MMM2 & 0.06 & \textbf{0.12} & 0 & 0 & 0\\
      10m11m & 0.04  &  \textbf{0.31}  & 0.05 & 0 & 0 \\
      Corridor & 0.03 & \textbf{0.29} & 0.1 & 0 & 0.09\\
    \bottomrule
\end{tabularx}
\end{table}
\begin{figure*}[ht]
    \centering
    \includegraphics[width=1\textwidth]{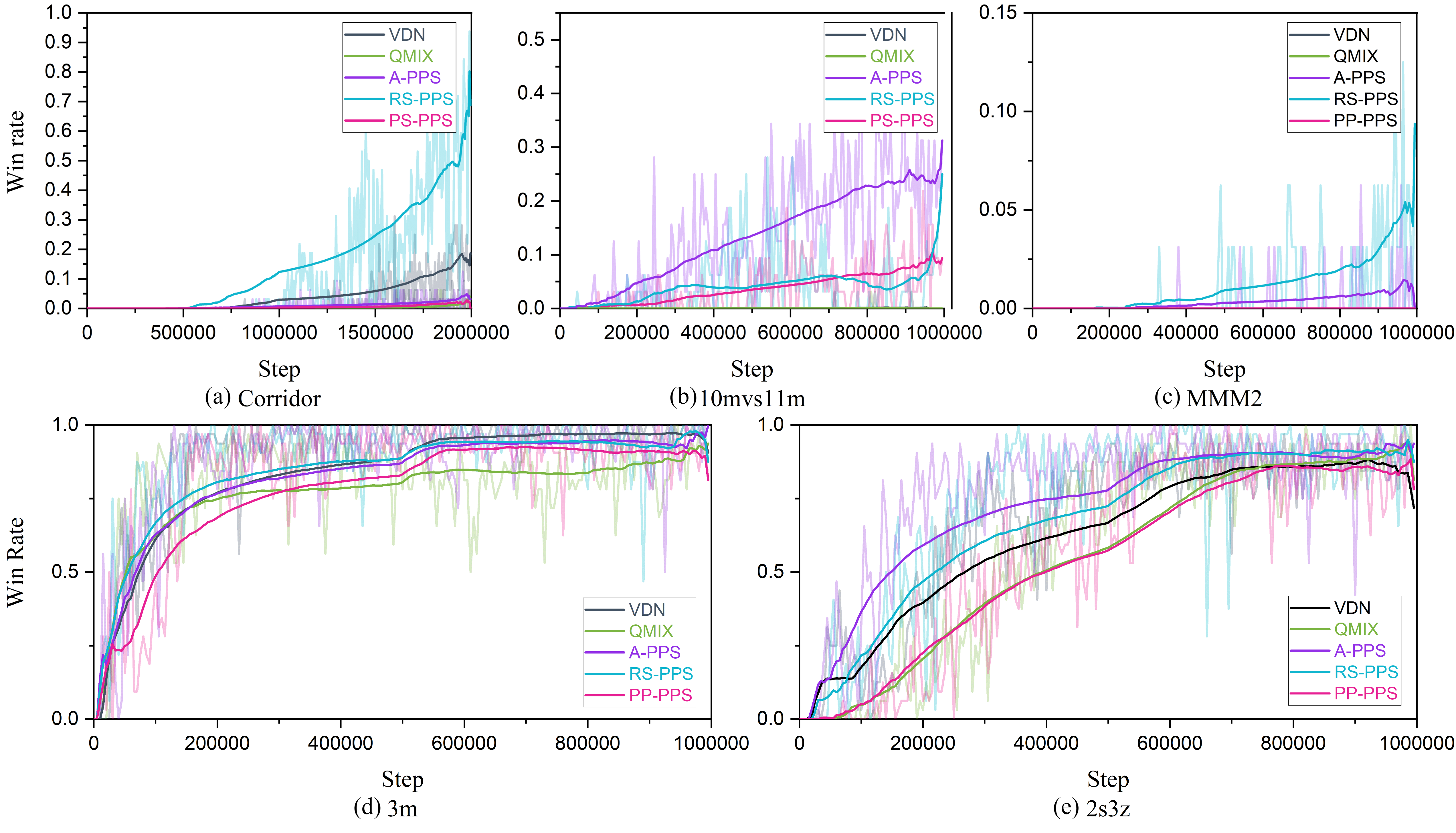}
        \caption{Comparative Performance(QMIX) in the SC2 environment. (a) (b) (c) are asymmetric environments and (d)(e) are symmetric environments. Performance of our method outperforms in (a) (b) (c) tasks compared with conventional approach.}
    \label{fig:result}
\end{figure*}
\begin{figure*}[ht]
    \centering
    \includegraphics[width=1\textwidth]{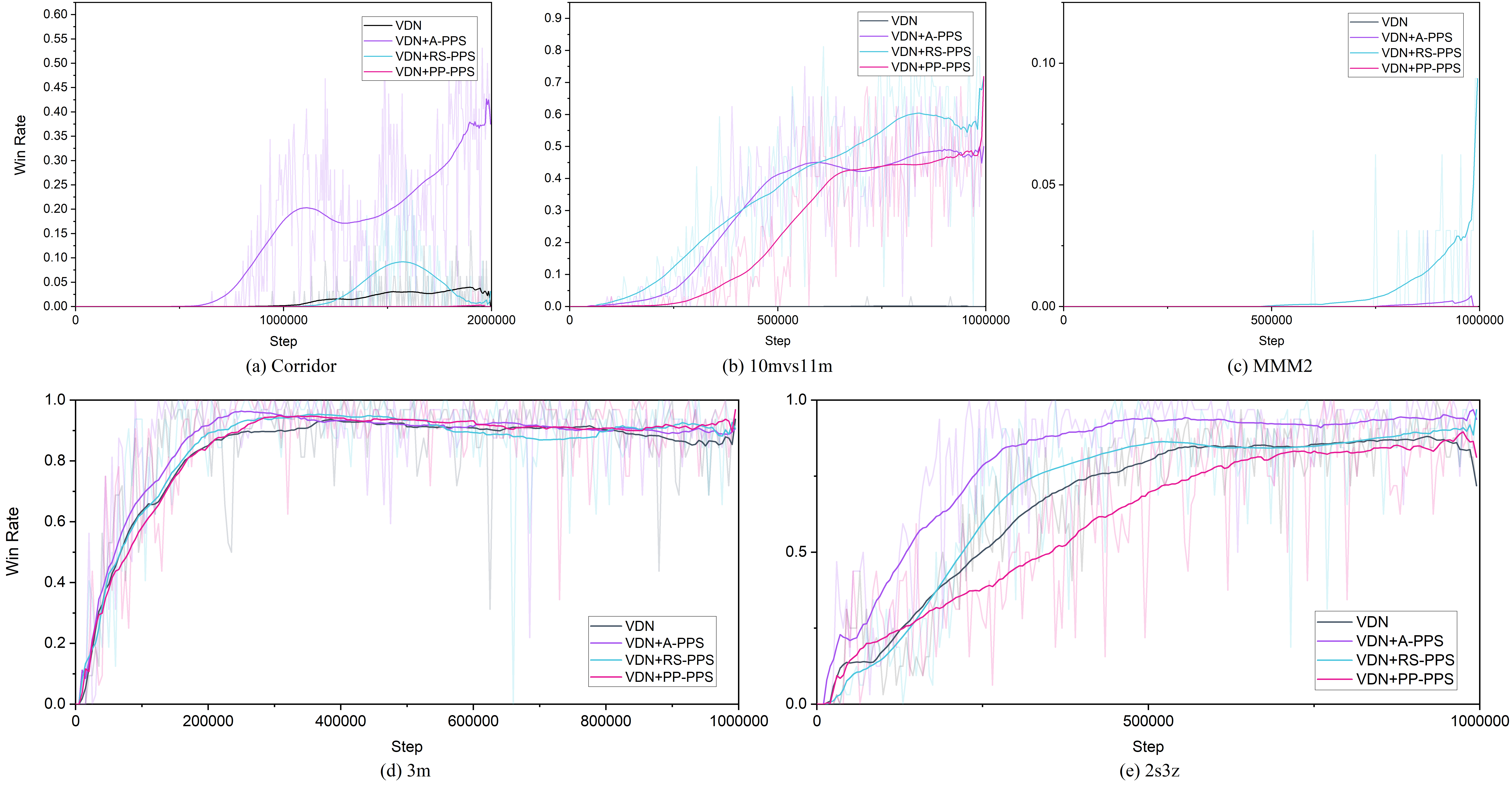}
        \caption{Comparative Performance(VDN) in the SC2 environment. (a) (b) (c) are asymmetric environments and (d)(e) are symmetric environments. Performance of our method outperforms in (a) (b) (c) tasks compared with conventional approach.}
    \label{fig:result}
\end{figure*}
In the A-PPS method, the average aggregation of the value network of each agent achieved better performance and win rate. Due to the average distribution of experience and trajectory, A-PPS can effectively accelerate training without significantly improving the effectiveness.
\par
To verify  robustness of our method, we further take ablation in tasks MMM2 (which stages for 1 Medivac, 2 Marauders versus 7 Marines 1 Medivac, 3 Marauders and 8 Marines), 10mvs11m (which stages for 10 Marins versus 11 Marins, for a total of 22 entities), and Corridor (which stages for 6 Zealots versus 24 Zergling, for a total of 30 entities) with a large number of agents in an asymmetric environment.
\par
Compared to A-PPS, RS-PPS in these experiments achieved more outstanding results. During training process, In contrast with QMIX and VDN, our proposed method achieved significant improvement and effectiveness in MMM2, 10m vs 11m, and Corridor tasks. In asymmetric tasks, due to the presence of other types and quantities of agents, the observed state space of each agent fluctuates greatly and increase explosively. Therefore, prior A-PPS can lead to a large number of erroneous experiences and thus RS-PPS, a reward buffer and reward scalability mechanism, is introduced into our model. The aggregation model based on reward buffer enables to correct the weights in aggregation process of models through the rewards obtained in the interval. Because correct decision made by each agent when facing exploration process will increase the weight occupied by the model in the aggregation process, that is, agent that allows other agents to learn value functions from largest accumulated reward function. So our RS-PPS method accelerates model convergence better when facing high-dimensional observation spaces and guides agents to make the right choices when facing complex observation states.
\par
As for PP-PPS, due to harness of fixed personalized points in the training process of PP-PPS, which means that fixed number of layers in the network model remains unchanged during aggregation process. However, in our experiments, for different agents, their corresponding representation layers are different, but fixed personalized point approach leads to different agents using the same number of representation layers. Therefore, PP-PPS leads to increasing deviation in representation of different agents, but compared to conventional methods, PP-PPS modified algorithm also introduce limited new knowledge during sharing process. Therefore, although PP-PPS performs poorly in multiple tasks compared to A-PPS and RS-PPS. However, experiments have still shown that PP-PPS method performs better than QMIX.

\section{Conclusion and Further Work}
In this paper, we consider a problem of slow training speed caused by uneven experience distribution of a single agent during multi-agent reinforcement learning, and the negative impact from incorrect actions of other agents on correct actions of agents in the team. Inspired by the federated learning algorithm to solve the Non-IId problem, three methods, A-PPS, RS-PPS, and PP-PPS, were introduced to solve these issues. We tested the three methods we proposed based on QMIX model in symmetric and asymmetric tasks of SMAC environment. The training convergence speed of A-PPS is faster than that of the basic QMIX model in the case of symmetric confrontation between a small number of agents. Compared with A-PPS and RS-PPS, in complex multi-agent circumstances, RS-PPS can achieve better victory rates and achieve tasks that cannot be completed by QMIX and VDN methods. Nevertheless, PP-PPS performs poorly in multiple tasks due to its fixed personalized representation layer, but its experimental results are better than QMIX. 
\par
Conclusively, proposed approaches, A-PPS, RS-PPS, and PP-PPS, can effectively accelerate convergence rate of multi-agent reinforcement learning training process and enable agents to complete tasks that were previously impossible to complete by QMIX and VDN. Our approaches enable to be applied to multi-agent reinforcement learning algorithms and produce effective results. Furthermore, we wll modifies personalized layer, PP-PPS algorithm, and explore novel architecture of feature representation so that intelligent agents can adaptively select effective feature representations instead of manually selecting fixed network layers. 

\begin{acks}
This research was supported by the Science Foundation of China University of Petroleum, Beijing (No. 2462020YXZZ024)
\end{acks}
\newpage



\bibliographystyle{ACM-Reference-Format} 
\bibliography{sample}


\end{document}